# Post Triangular Rewiring Method for Shorter RRT Robot Path Planning

**Jin-Gu Kang, Jin-Woo Jung** *

Department of Computer Science and Engineering, Dongguk University, Seoul 04620, Korea

**Abstract**

This paper proposed the 'Post Triangular Rewiring' method that minimizes the sacrifice of planning time and overcomes the limit of Optimality of sampling-based algorithm such as Rapidly-exploring Random Tree (RRT) algorithm. The proposed 'Post Triangular Rewiring' method creates a closer to the optimal path than RRT algorithm before application through the triangular inequality principle. The experiments were conducted to verify a performance of the proposed method. When the method proposed in this paper are applied to the RRT algorithm, the Optimality efficiency increase compared to the planning time.

## 1. Introduction

The path planning [1] is to create or plan a path that a mobile robot can efficiently move from a starting point to a destination point in Euclidean space, avoiding obstacles, with Optimality, Clearance, and Completeness. Of these, Optimality means always ensuring to plan a path with the optimal path length, Clearance indicates about how low the probability of collision between obstacles and the mobile robot. and Completeness indicates if that a path can always be planned in the presence of a solution (if a path can be created from the starting point to the destination point without colliding with obstacles).

This paper proposed a sampling-based RRT (Rapidly-exploring Random Tree) algorithm [2] that does not guarantee Optimality is treated. The RRT algorithm can be summarized as a method of planning a path by repeating the act of inserting a randomly sampled position as a child node in a tree with the starting point as the root node until reaching the destination point. With this algorithm, the tree extends out in the shape of a stochastic fractal and has a process of finding the destination point.

The sampling-based algorithm [3-4], including the RRT algorithm, has the advantage of being able to plan a path in less time with less computation than the algorithms of the classical path planning algorithm such as Visibility Graph-based [5], Cell Decomposition-based [6], and Potential Field-based [7]. On the other hand, it does not guarantee Optimality, and has the disadvantage that Completeness is guaranteed probabilistically. The latter is also called Probabilistic completeness [8], which means that Completeness is guaranteed when the number of random samples is infinite but may not be guaranteed when the number of random samples is finite. The purpose of this paper is to study the path planning algorithm of the RRT algorithm, which guarantees Completeness and shows performance closer to Optimality than the related works.

The proposed 'Post Triangular Rewiring' method is effective in path planning algorithms that do not guarantee Optimality like the RRT algorithm and show locally piecewise linear shape and can be applied as a post-processing method after a path is planned through such an algorithm.

Of course, since this proposed method have the greatest advantage of the sampling-based path planning algorithm is the fast planning speed due to the small amount of computation compared to the classical path planning algorithms [3], the amount of computation added by the proposed method should not be large compared to the RRT algorithm.

To verify the performance of the method proposed in this paper, the performance verification through mathematical modeling. And the planning time and the path length of the First Complete Path(The path first reaches a destination point form a starting point) is compared and analyzed through simulation when the method proposed in various environments is applied to the RRT algorithm and related works and when not applied.

## 2. Rapidly-exploring Random Tree (RRT)

The RRT algorithm is a representative algorithm of the sampling-based path planning algorithm and was proposed by Steven M. LaValle in 1998 [2]. It is useful for planning a path considering the conditions of non-holonomic constraints and is designed to have high degrees of freedom.

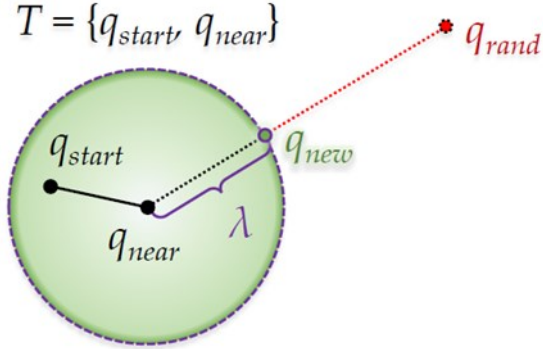

Figure 1. The process of the RRT algorithm: In the case of creating a new node $q_{new}$ at a position separated by a step length $\lambda$ in the direction of the random sample position($q_{rand}$) based on the $q_{near}$ node (position) nearest to the random sample position ($q_{rand}$) in the tree $T$ with the starting point $q_{start}$ as the root node.

When a random sample is generated in the Configuration space, as shown in Figure 1, the nearest node to the position of the random sample is found among the nodes constituting the tree with the starting point as the root node. A new node is created at a position away from the node by a step length in the direction of the random sample position and inserted into the tree. If the random sample position is closer than the step length, a new node is created at the random sample position and inserted into the tree. This tree extension process is repeated until the destination point is reached.

## 3. Proposed Post Triangular Rewiring Method

The proposed 'Post Triangular Rewiring' method can be applied to path planning algorithms that do not guarantee Optimality such as RRT (Rapidly-exploring Random Tree) algorithm, rewire based on the triangular inequality principle [4].

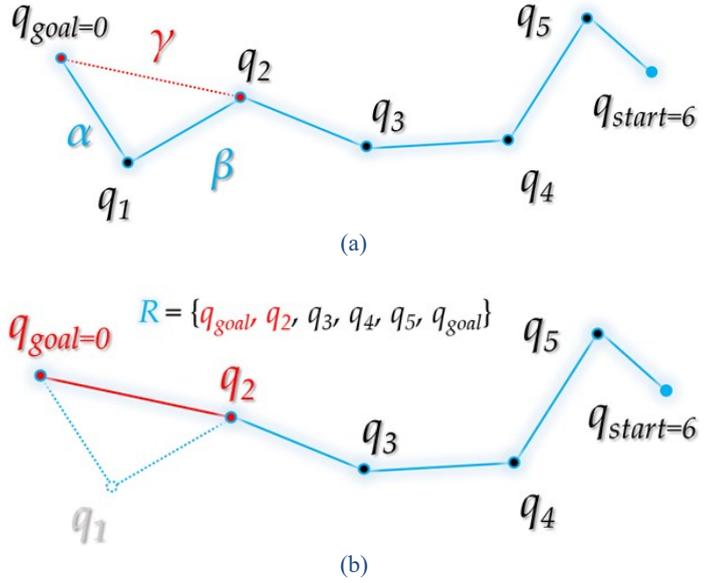

Figure 2. Summary of Post Triangular Rewiring method: (a) When the line segment $\gamma$ with node $q_0$ and its grandparent node $q_2$ in tree $R$ is free from obstacle collision (Distance: $\gamma<\alpha+\beta$); (b) The grandparent node $q_2$ of the node $q_0$ is connected as the parent node of the node $q_0$, and the parent node $q_1$ is deleted from the tree.

It is a post-processing method after a path is planned through such an algorithm. As shown in Figure 2(a), when there is no obstacle between the current focusing node and its grandparent node (Collision free), a shorter path can be created based on the triangular inequality as shown in Figure 2(b). The focusing node and the grandparent node of the node are connected as a parent node, and rewire such as deleting the parent node. Therefore, when the proposed Post Triangular Rewiring method is applied, it is possible to modify the path closer to the optimum than the original path of the RRT algorithm.

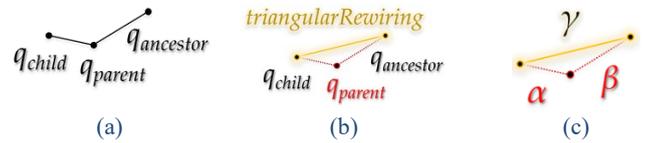

Figure 3. (a) A part of a tree consisting of a node $q_{child}$, its parent node $q_{parent}$ and the parent node $q_{ancestor}$ of $q_{parent}$; (b) If there is no obstacle between $q_{child}$ and $q_{ancestor}$, delete the edge connecting to $q_{parent}$ in each node; (c) The edge $\alpha$ connecting $q_{child}$ and $q_{parent}$, the edge $\beta$ connecting $q_{parent}$ and $q_{ancestor}$, and the edge $\gamma$ connecting $q_{child}$ and $q_{ancestor}$.

Figure 3 (a) and (b) shows an example of rewiring [4], which can be expressed as Equation 1:

$$q_{ancestor} = \xi(q_{parent}) = \xi^2(q_{child}). \qquad (1)$$

Here, $\xi()$ is a function that receives the node as a variable and returns the parent node of that node. The n-squared ($n\geq0$) of the $\xi()$ function can be expressed as $\xi^n(q_i) := (\overbrace{\xi \circ \xi \circ ... \circ \xi}^{n})(q_i)$, and if $n$ is 0, $\xi^0(q_i) := q_i$ holds. That is, $m_F(q_i,k)$ ($k>0$) becomes the midpoint between $m_F(q_i,k-1)$ and $\xi(q_i)$. At this time, $d$ becomes $(d_k-1)/2$.

Also, by substituting into the triangular inequality in Equation 1, it can be expressed as in Figure 3(c) and Equation 2:

$$\alpha + \beta \geq \gamma, \qquad (2)$$

Equation 3 below shows the distance relation between ancestor nodes based on $q_{child}$:

$$D(q_{child}, \xi(q_{child})) + D(\xi(q_{child}), \xi^2(q_{child})) \geq D(q_{child}, \xi^2(q_{child})). \quad (3)$$

Equation 3 is the distance $D()$ expressed in Equation 1 between the $n$-th ancestor nodes based on $q_{child}$ in Equation 2, which represents the distance between each node as a triangular inequality, and then substituted. Here, $D()$ means the distance between two points, That is, Equations 1-3 locally shows that the path becomes shorter than when rewiring.

The following Algorithms 1-2 shows a pseudocode of the proposed Post Triangular Rewiring method.

**Algorithm 1.** Pseudocode of the Proposed 'Post Triangular Rewiring' Method.

**Input:**
$R \leftarrow$ path from $\{RRT / ...\}$
$C \leftarrow$ position set of all (measured) boundary points in all (known) obstacles
$\varepsilon \leftarrow$ threshold value of minimum Clearance
**Output:**
$R \leftarrow$ modified path $R$
**Initialize:**
$f_{modify} \leftarrow true$

**Procedure** postTriangularRewire
**Begin**
1   **While** $f_{modify}$ **Do**
2       $f_{modify} \leftarrow false$   // is the path modified
3       $t \leftarrow 0$   // index of the currently focused point
4       $q_{child} \leftarrow$ first point in $R$
5       $q_{parent} \leftarrow$ next point of $q_{child}$ in $R$
6       **While not** [$q_{parent}$ is last point in $R$] **Do**
7           $q_{ancestor} \leftarrow$ next point of $q_{parent}$ in $R$
8           **If not** isTrapped($q_{child}$, $q_{ancestor}$, $C$) **Then**
9               $R \leftarrow$ rewire($R$, $\varepsilon$, $t$, $f_{modify}$)
10          **Else**
11              $t \leftarrow t + 1$
12              $q_{child} \leftarrow t$-th point in $R$
13              $q_{parent} \leftarrow$ next point of $q_{child}$ in $R$
**End**

The input value of the Post Triangular Rewiring method consists of the path $R$ planned through the path planning algorithm such as the RRT algorithm, the obstacle area information $C$, and the threshold value $\varepsilon$ of the minimum Clearance.

$f_{modify}$ is a variable that determines whether the input path $R$ has been modified by this method, and if the path is modified even once, the entire process is repeated. If the path modification does not occur in the process of repeating again, the algorithm is terminated. $t$ refers to the index of the waypoint of $R$ that is currently focused. That is, if $t$ is 0, it means the starting point, which is the first point of $R$.

In $R$, when the first focusing point is $q_{child}$, the next point of that point $q_{child}$ is $q_{parent}$, and the next point of that point $q_{parent}$ is mentioned $q_{ancestor}$, it is determined whether the distance between $q_{child}$ and $q_{ancestor}$ is free from obstacle collision (isTrapped() function). If it is free from collision, it calls rewire(). rewire() connects $q_{child}$ and $q_{ancestor}$, and the $q_{parent}$ between them is deleted from the path. If $R$ and $t$ are updated due to rewire(), update $q_{child}$ (the $t$-th waypoint of $R$), $q_{parent}$ and $q_{ancestor}$ accordingly. If $q_{parent}$ is the last point in $R$, check $f_{modify}$. Otherwise, repeat the above process again for the updated $q_{child}$ and $q_{ancestor}$.

Here, path modification by rewire() deletes the waypoints and makes a closer to optimal path.

**Algorithm 2.** Pseudocode of the 'Rewire' Function from the Proposed Method.

**Input:**
$R \leftarrow$ path $R$ from postTriangularRewire
$t \leftarrow$ point index $t$ from postTriangularRewire
$f_{modify} \leftarrow$ boolean $f_{modify}$ from postTriangularRewire
**Output:**
$R \leftarrow$ modified path $R$
$f_{modify} \leftarrow$ result of boolean $f_{modify}$   // return by reference

**Procedure** rewire **From** postTriangularRewire
**Begin**
1   $q_{child} \leftarrow t$-th point in $R$
2   $q_{parent} \leftarrow$ next point of $q_{child}$ in $R$
3   $q_{ancestor} \leftarrow$ next point of $q_{parent}$ in $R$
4   $R \leftarrow$ **Delete** path<$q_{child}$, $q_{parent}$> and path<$q_{parent}$, $q_{ancestor}$> from $R$
5   $R \leftarrow$ **Insert** path<$q_{child}$, $q_{ancestor}$> to $R$
6   $f_{modify} \leftarrow true$
**End**

The input value of rewire() of the Post Triangular Rewiring method consists of the path $R$, the focusing point index $t$, and the path modification $f_{modify}$ from the Post Triangular Rewiring method. Rewiring is performed on the $t$-th waypoint $q_{child}$ of $R$, the next point $q_{parent}$, and the next point $q_{ancestor}$ of that point again. First, delete the path between $q_{child}$ and $q_{parent}$, and the path between $q_{parent}$ and $q_{ancestor}$. Then insert the path between $q_{child}$ and $q_{ancestor}$. Finally, $f_{modify}$ returns 'true' because the path has been modified.

The following Figure 4 shows the overall flow chart of the proposed Post Triangular Rewiring method. Here, $\xi^t(q_{goal})$ means the $t$-th next waypoint from the starting point $q_{goal}$ of the path $R$, and $\xi^{t+n}(q_{goal})$ means the $n$-th next waypoint in the $\xi^t(q_{goal})$. That is, there are $n$ waypoints between $\xi^t(q_{goal})$ and $\xi^{t+n}(q_{goal})$.

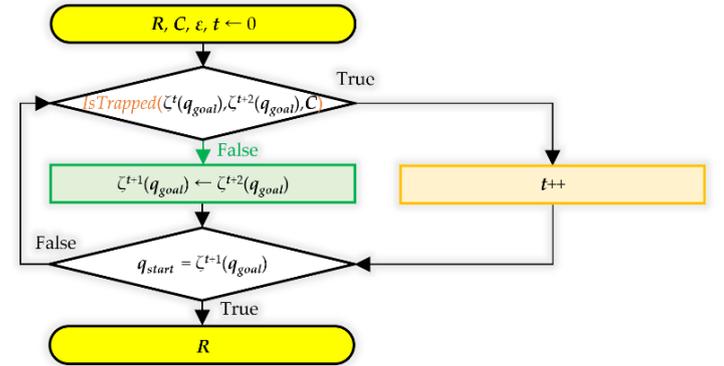

Figure 4. Flow chart of the proposed Post Triangular Rewiring method.

## 4. Experimental Results

To verify the performance of the Post Triangular Rewiring method proposed in this paper, the path between RRT (Rapidly-exploring Random Tree) in various environments through simulation and the RRT algorithm to which the proposed method is applied Path planning results were compared.

The compared performance measures are the average of all trials when each algorithm is repeated 100 times (sampling position is changed for each trial), the path length(px) and the planning time(ms), of the first complete path(From starting point to destination point until the first path is planned).

The following Figure 8 shows the four environment maps used in the experiment. Here, the green circle (S) refers to the starting point, and the purple circle (G) refers to the destination point. And a black polygon with a yellow border (blue in the experimental results) means an obstacle. The size of all environment maps is 600*600px, and the step length is 30px.

In the related works of the path planning algorithm, various environmental maps were considered and utilized to confirm the performance of the proposed method. It is important which environment map to use because the efficiency of the performance measures expected during the experiment is somewhat different depending on the composition, such as the number, arrangement, or shape of obstacles. In this paper, the four environmental maps shown in Figures 5-8 are used to verify the performance of the proposed method. These maps are part of the experimental environment proposed by Jihee Han [9] in 2017, and the efficiency of the following characteristics and performance measures is expected for each map.

Map 1 of Figure 5 seems to be an environment that is efficiency to verify for Optimality and Completeness, and it is an environment unfavorable to sampling-based path planning algorithms such as the RRT algorithm. Because the probability of finding a solution is low, many samplings are required. Map 2 of Figure 6 is an environment that is efficiency to verify for the Optimality and Completeness of the path planning algorithm. Map 3 of Figure 7 is an environment that is efficiency to verify the Optimality of the path planning algorithm and the planning time as it is composed of obstacles (50 squares) that are close to curved. The costly classical path planning algorithm is unfavorable. Map 4 of Figure 8 is an environment that is efficiency to verify for the Completeness and planning time of the path planning algorithm, and is an environment unfavorable to sampling-based path planning algorithms such as the RRT algorithm.

Since the sampling-based path planning algorithm depend on Probabilistic completeness, the number of samples and planning time required becomes extremely large and long as there are narrowly or few entrances in the direction to the destination point.

Table 1. Computer performance for simulation.

| H/W | Specification |
|---|---|
| CPU | Intel Core i7-6700k 4.00GHz (8 CPUs) |
| RAM | 32768MB (32GB DDR4) |

Table 1 shows the performance of the computer used in the simulation. The simulator used for the simulation [4] was developed based on C# WPF (Microsoft Visual Studio Community 2019 Version 16.1.6 Microsoft .NET Framework Version 4.8.03752), and only a single thread was used for calculations except for the visual part. Of course, there may be differences in planning time during simulation depending on computer performance. Therefore, in the experiment of this paper, the planning time is not compared absolutely, but relatively based on the RRT algorithm.

Analyze the experimental results (path length, planning time) when the Post Triangular Rewiring method is applied to the RRT algorithm, and its path planning results in the four environment maps presented in the experimental environment.

In each map, a figure of the path planning (in case of a single trial) result for each algorithm is shown, and the results of an experiment (when it is repeated) on the performance measures are shown in numerical values in each table (the figure for each algorithm is not the result of repeated trials). As for the one of the repeated trials, there may be a large difference between the performance observed with the visually and the numerical results shown in the table). The figure is shown to refer to the shape of the planned path for each algorithm. In addition, it is determined whether there is a section in which the piecewise linear shape path is smoothed by the proposed Post Triangular Rewiring method.

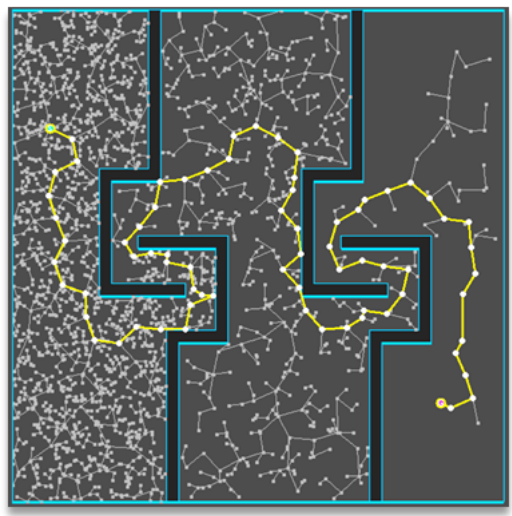

(a)

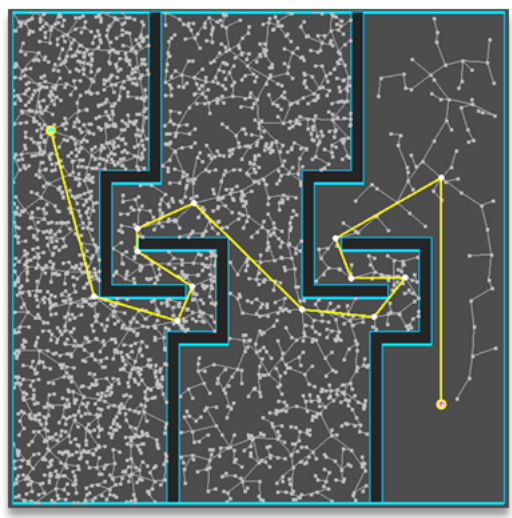

(b)

Figure 5. Experimental results of map 1: (a) RRT; (b) Proposed method applied.

Figure 5 shows the path planning results for map 1 among the presented environment maps for each algorithm. In visually, a path of the Post Triangular Rewiring method seems to be the shortest.

Table 2. Experimental results of map 1 <The parentheses on the right of each number (average of repeated 100 times) are relative ratios based on 100% RRT (values less than 1 are counted as 1)>.

| Performance | RRT | Proposal |
|---|---|---|
| Path length(px) | 1932 *(100%)* | 1403 *(72%)* |
| Planning time(ms) | 687 *(100%)* | 688 *(100%)* |

Table 2 shows the path planning results for map 1 (average when repeated 100 times) among the presented environment maps for each algorithm. The path length applying the Post Triangular Rewiring method becomes 72% (1403/1932(%)) compared to the RRT, which is the shortest, and the planning time is all similar.

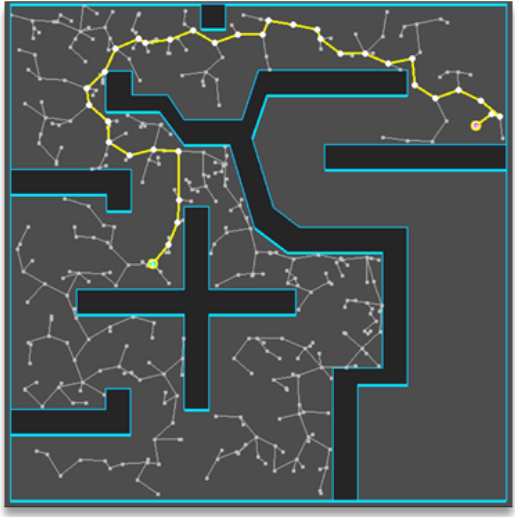

(a)

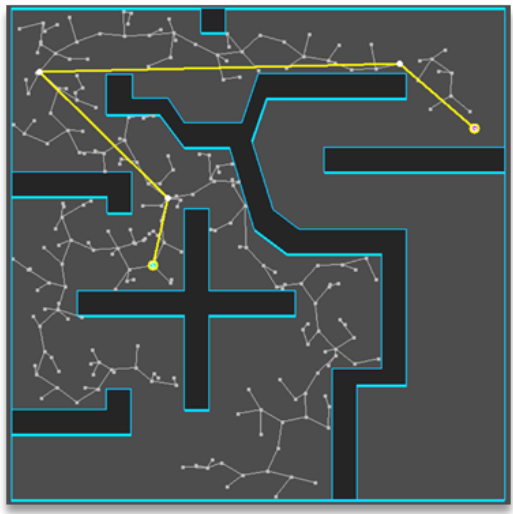

(b)

Figure 6. Experimental results of map 2: (a) RRT; (b) Proposed method applied.

Figure 6 shows the path planning results for map 2 among the presented environment maps for each algorithm. In visually, a path of the Post Triangular Rewiring method seems to be the shortest.

Table 3. Experimental results of map 2 <The parentheses on the right of each number (average of repeated 100 times) are relative ratios based on 100% RRT (values less than 1 are counted as 1)>.

| *Performance* | RRT | Proposal |
|---|---|---|
| **Path length(px)** | 969 *(100%)* | 799 *(82%)* |
| **Planning time(ms)** | 10 *(100%)* | 11 *(100%)* |

Table 3 shows the path planning results for Map 2 (average when repeated 100 times) among the presented environment maps for each algorithm. The path length applying the Post Triangular Rewiring method becomes 82% (799/969(%)) compared to the RRT, which is the shortest, and the planning time is all similar.

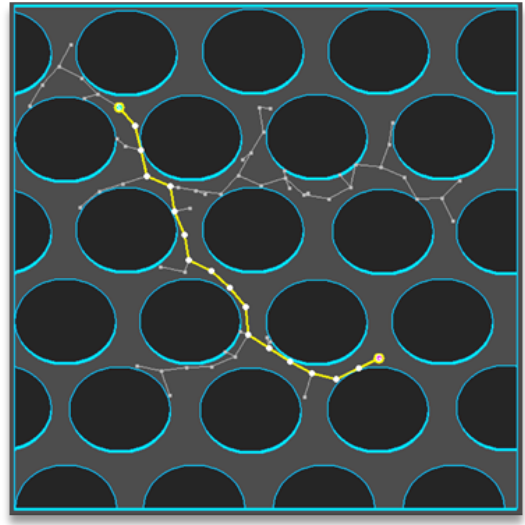

(a)

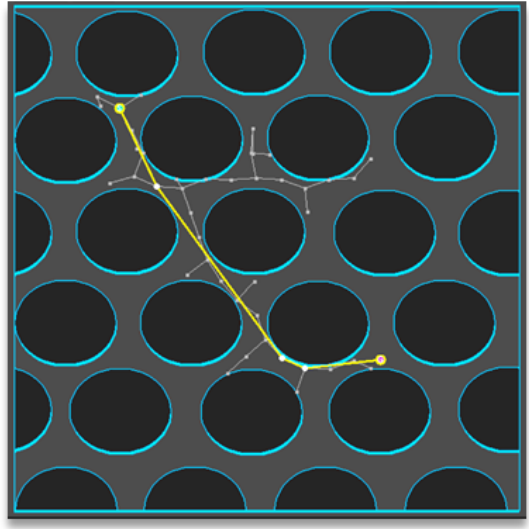

(b)

Figure 7. Experimental results of map 3: (a) RRT; (b) Proposed method applied.

Figure 7 shows the path planning results for map 3 among the presented environment maps for each algorithm. In visually, a path of the Post Triangular Rewiring method seems to be the shortest.

Table 4. Experimental results of map 3 <The parentheses on the right of each number (average of repeated 100 times) are relative ratios based on 100% RRT (values less than 1 are counted as 1)>.

| *Performance* | RRT | Proposal |
|---|---|---|
| **Path length(px)** | 591 *(100%)* | 529 *(89%)* |
| **Planning time(ms)** | 6 *(100%)* | 7 *(100%)* |

Table 4 shows the path planning results for map 3 (average when repeated 100 times) among the presented environment maps for each algorithm. The path length applying the Post Triangular Rewiring method becomes 89% (529/591(%)) compared to the RRT, which is the shortest, and the planning time is all similar.

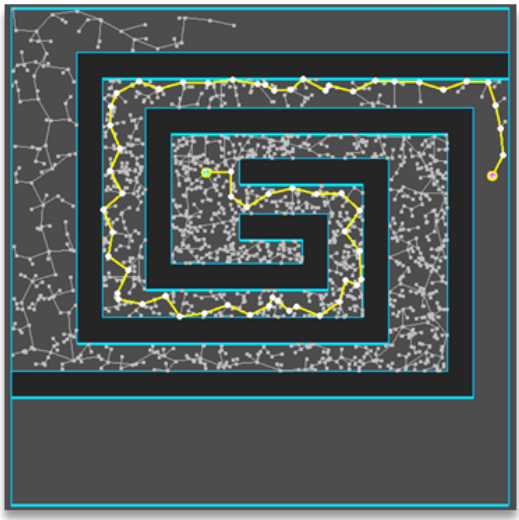

(a)

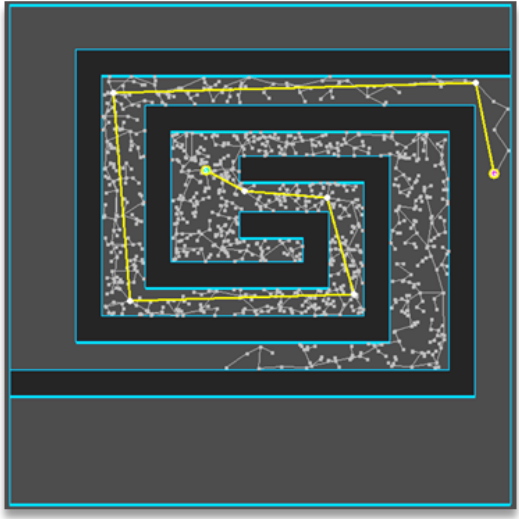

(b)

Figure 8. Experimental results of map 4: (a) RRT; (b) Proposed method applied.

Figure 8 shows the path planning results for map 4 among the presented environment maps for each algorithm. In visually, a path of the Post Triangular Rewiring method seems to be the shortest.

Table 5. Experimental results of map 4 <The parentheses on the right of each number (average of repeated 100 times) are relative ratios based on 100% RRT (values less than 1 are counted as 1)>.

| Performance | RRT | Proposal |
|---|---|---|
| Path length(px) | 1533 (100%) | 1304 (85%) |
| Planning time(ms) | 1526 (100%) | 1527 (100%) |

Table 5 shows the path planning results for map 4 (average when repeated 100 times) among the presented environment maps for each algorithm. The path length applying the Post Triangular Rewiring method becomes 85% (799/969(%)) compared to the RRT, which is the shortest, and the planning time is all similar.

Overall, the application of the Post Triangular Rewiring method showed good performance in the path length for all maps, showing that the proposed method is effective in terms of Optimality.

## 5. Conclusion

In this paper, we propose a Post Triangular Rewiring method that can minimize the sacrifice of planning time and overcome the limitations of the optimization of the sampling derivation algorithm.

The proposed Post Triangular Rewiring method makes a more optimal path. In addition, as a post-processing method, it has the advantage of being applicable to all path planning algorithms that plan a locally piecewise linear path. Simulations were performed to verify the performance of the RRT algorithm to which the Post Triangular Rewiring method was applied. It was verified that the path length was shortened by 11-28% (average 18%) when applied to the RRT algorithm in the four different environment maps. As a result, the RRT algorithm applying the proposed Post Triangular Rewiring method showed a more optimal path.

## Conflict of Interest

No potential conflict of interest relevant to this article was reported.

## Acknowledgement

This work was supported by the National Research Foundation of Korea(NRF) grant funded by the Korea government(MSIT) (No. 2020R1F1A1074974), the Ministry of Trade, Industry and Energy(MOTIE) and Korea Institute for Advancement of Technology(KIAT) through the International Cooperative R&D program. (Project No. P0016096), the AURI(Korea Association of University, Research institute and Industry) grant funded by the Korea Government(MSS : Ministry of SMEs and Startups). (No.S3041234, HRD program for Enterprise linkages R&D) and the MSIT(Ministry of Science and ICT), Korea, under the ITRC(Information Technology Research Center) support program(IITP-2021-2020-0-01789) supervised by the IITP(Institute for Information & Communications Technology Planning & Evaluation).

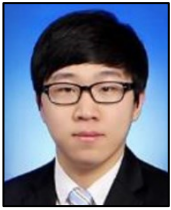
**Jin-Gu Kang** received the B.S. and LL.B. degrees from Dongguk University at Seoul, Korea in 2019, and received the M.S. degree in Computer Science and Engineering from Dongguk University at Seoul, Korea in 2021. His research areas include robot path planning, wireless MAC protocols and deep learning-based image processing.
E-mail: kanggu12@dongguk.edu

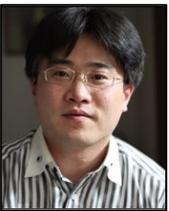
**Jin-Woo Jung** received the B.S. and M.S. degrees in Electrical Engineering from Korea Advanced Institute of Science and Technology (KAIST), Korea, in 1997 and 1999, respectively and received the Ph.D. degree in Electrical Engineering and Computer Science from KAIST, Korea in 2004. Since 2006, he has been with the Department of Computer Science and Engineering at Dongguk University at Seoul, Korea, where he is currently a Professor. His current research interests include human behavior recognition, mobile robot and intelligent human-robot interaction.
E-mail: jwjung@dongguk.edu